# Efficient Analysis of Complex Diagrams using Constraint-Based Parsing[1,2]

Robert P. Futrelle and Nikos Nikolakis

Biological Knowledge Laboratory, College of Computer Science
161 Cullinane Hall, Northeastern University, Boston, MA 02115,
{futrelle,nikos}@ccs.neu.edu

cmp-lg/9505015    5 May 1995

**Abstract**

*This paper describes substantial advances in the analysis (parsing) of diagrams using constraint grammars. The addition of set types to the grammar and spatial indexing of the data make it possible to efficiently parse real diagrams of substantial complexity. The system is probably the first to demonstrate efficient diagram parsing using grammars that easily be retargeted to other domains. The work assumes that the diagrams are available as a flat collection of graphics primitives: lines, polygons, circles, Bezier curves and text. This is appropriate for future electronic documents or for vectorized diagrams converted from scanned images. The classes of diagrams that we have analyzed include x,y data graphs and genetic diagrams drawn from the biological literature, as well as finite state automata diagrams (states and arcs). As an example, parsing a four-part data graph composed of 133 primitives required 35 sec using Macintosh Common Lisp on a Macintosh Quadra 700.*

**Introduction**

Future electronic documents will be enhanced by graphics that are represented as structured objects, rather than bitmaps. Diagrams of anatomy or cells in textbooks could be accessed by their components; maps could be accessed by reference to specific buildings or streets, etc. Incorporating structured graphics into future information systems will require progress on many fronts, including systems for analyzing existing graphics, knowledge-based tools for creating graphics, and intelligent tools for retrieving and interacting with structured graphics.

There are specialized systems that can efficiently analyze complex diagrams (one hundred or more instances of primitives). But it has been difficult to adapt them to domains other than what they were designed for. At the other end of the spectrum, there are approaches to visual parsing that use general grammatical models and are thus adaptable. But these have not been efficient enough in practice to analyze complex diagrams. Our approach is both efficient and adaptable. Many of these systems have been applied to domains such as engineering drawings and circuit diagrams. But the published technical literature contains far more diagrams. For example, the biological literature that we focus on publishes about 2.5 million diagrams per year, mostly data graphs and gene diagrams.

Descriptions of certain aspects of our system have been published [1, 2]. In this paper, we first explain our constraint grammars and show an example of a complex data graph the system can parse. Then a small grammar is given and the parsing process is explained in detail. Spatial indexing, the key to much of the system's efficiency, is described. Some aspects of the large grammar used to parse the data graph are discussed (the grammar is presented in the Appendix). Gene diagrams and finite-state automata diagrams the system has parsed are shown. Finally, we discuss the relation of this work to other methods.

**Constraint grammars and efficient parsing**

Graphics constraint grammars are particularly useful in diagram analysis [2, 3]. In these grammars a rule consists of a production with a left-hand-side symbol (LHS), one or more right-hand-side (RHS) constituents, and a body. In our grammar there are two rule types, ordinary rules and set rules. Some constituents are primitive graphical objects such as lines, polygons, circles, Bezier curves and text; others may reference LHS symbols, which are higher level objects. The body of the rule contains constraints on constituents, including geometric relations among them, on set members, and on sets as a whole. Constraints can refer to a variety of geometric properties such as position, shape, length, size, components (e.g., endpoints of a line, center of an object), etc. More powerful constraints operate between objects, such as nearness or relative position (*above*, *below*). The most powerful constraints are those that operate across entire sets of objects, such as requiring that all the objects in a set be horizontally aligned, or connected. These latter relations allow the parser to rapidly collect together large sets of related items, reducing the effective size of the problem. Although many subsets of a set may satisfy a relation, our algorithm chooses the maximal set. A precomputed spatial index is used to make the computation of the constraints more efficient. Solutions of a rule are generated by finding tuples of constituents in the diagram that satisfy all the constraints in the body of a rule. At this point one or more LHS objects are created with those constituents. Each LHS object has full status as a graphical object, with region, bounding box, center, etc., so it can participate as a constituent in other rules.

---

[1] Work supported in part by grant from the Department of Energy, Award No. DE-FG02-93ER61718.

[2] To be published in the proceedings of ICDAR '95 (Third Intl. Conf. on Document Analysis and Recognition, Montreal, Canada, August 14-16)

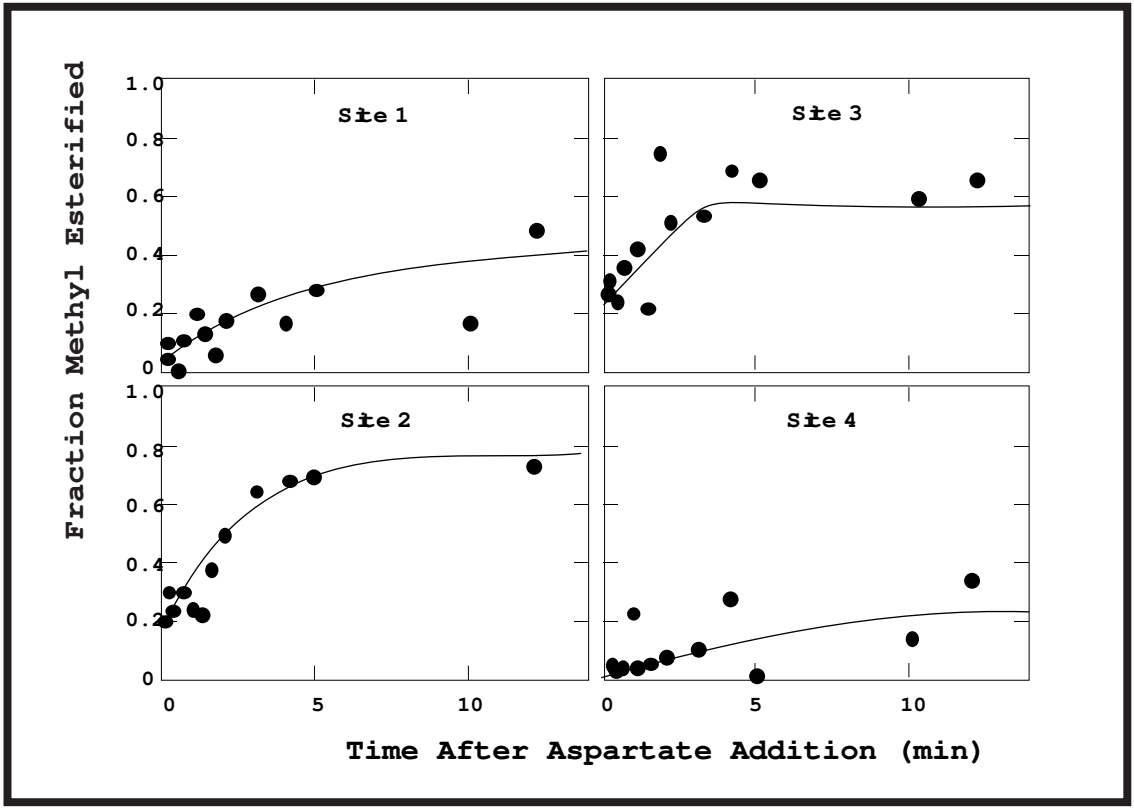

**Figure 1.** A four-part data graph consisting of 133 graphics primitives (lines, curves, circles, polygons and text) taken from [4]. Parsing the diagram, using the grammar G2 in the Appendix, required 35 sec on a Macintosh Quadra 700, running Macintosh Common Lisp 2.0. (An additional 86 sec was required to precompute the spatial indexes). The parse resulted in 4 solutions, one for each part, including the identification of the scale lines with tick marks and tick labels, the axis labels, the data points and data curves.

In our system, parsing proceeds top-down and depth-first. There is no ordering of constituents implied by the RHS; the body of the rule controls the order. A user can write grammar rules that lead to efficient parsing by specifying constraints that cut down on the number of elements that need to be examined or that are passed to lower rules. The solution strategy is more in the spirit of constraint satisfaction [5] than classical parsing — limited solution spaces are generated and then further restricted by the application of constraints.

**Example: A simple diagram and a small grammar**

In x,y data graphs, the long scale lines together with their short tick marks can involve a large number of lines and pieces of text. But these lines have a very regular organization, as shown in Fig. 1. When the *a-e* portions of the diagram in Fig. 2 (24 items) are analyzed for X-Ticks, a horizontal line with attached tick marks, only the two analyses $XT_1$ and $XT_2$ will result, according to the grammar G1, below. In particular,

- In *a* the two ticks on the far left are excluded because they don't touch the horizontal line.

- In *b*, the four lower ticks are aligned with one another, forming a set distinct from those in *a*.

- In *c* there are only two ticks, less than required.

- In *d* there are three vertical lines, which are too long.
- In *e* the ticks are not associated with a horizontal line.

The constraint grammar G1 for X-Ticks is:

Rule 1:
```
X-Ticks -> Ticks X-Line
           (X-Line)
           (Ticks (touch X-Line ?)
                :constraints
                    (> (number-of Ticks) 2));
```
Rule 2:
```
X-Line -> Line
          (:constraints
             (horizp Line) (long Line));
```
Rule 3:
```
Ticks -> Set (Line)
         (:element-constraints
             (vertp Line)  (short Line))
         (:constraints   (horiz-aligned));
```

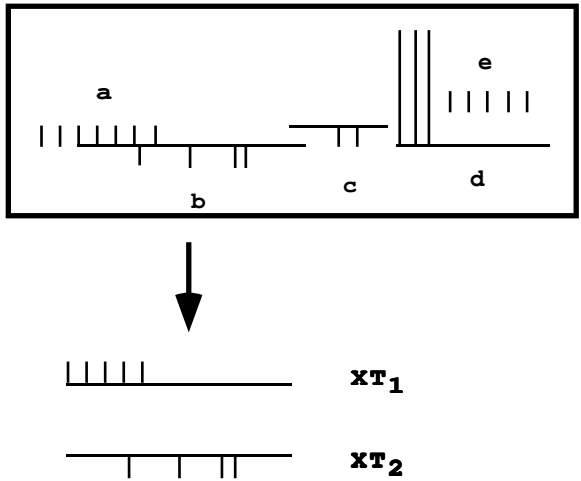

**Figure 2.** A diagram with 24 lines in the *a-e* portion which yields two X-Tick structures, $XT_1$ and $XT_2$, according to the grammar G1.

Grammar G1 illustrates our strategy:

- In Rule 1, "(X-Line)" appears first in the body, so it is processed first. It refers to X-Line in Rule 2 where it expands to the primitive, Line.
- A solution space for X-Line is generated by Rule 2 which consists of all lines which are horizontal and long. There are three such lines in Fig. 2, leading to three potential solutions. The "long" constraint is one related to the overall size of the diagram and to the distribution of line lengths[3].
- For each X-Line solution, the Ticks rule, Rule 3, is entered, inheriting the *context* attribute determined by the form "(touch X-Line '?)" in Rule 1. The value of *context* in this case, the value of "?", is all graphical objects which touch the given X-Line. Rule 3 states that every Ticks solution is a set, in this case a set of Lines. The Lines must be drawn from the set of objects in the context inherited by Ticks. The constraints on each member of the set are that they are vertical and short. The constraints on each set as a whole is that the elements are horizontally aligned with one another.
- The processing returns to Rule 1 where the set size constraint is imposed, eliminating the two-tick structure *c*. The top-level Rule 1 is complete, giving two solutions $XT_1$ and $XT_2$.

The analysis is efficient because of the continued restriction of the context as it is passed down the search tree. The set intersections needed for this are performed with a linear time algorithm using tagged objects. The pre-computed spatial index is used to achieve substantial speedup in computing a wide variety of geometric relations and constraints. The analysis of Fig. 2 (N=24) required 0.28 sec. to parse and return the two solutions plus 9 seconds to build the spatial index.

The approach just described integrates a number of techniques that make it easy to write grammars that describe a wide variety of types of complex diagrams. But at the same time, parsing is efficient. The success of the approach rests on a number of factors:

- Matching all aspects of the system to the spatial organization that people perceive in diagrams and use in drawing diagrams.
- Using sets as a fundamental component of the grammatical formalism.
- Using equivalence relations (*near, aligned*) to partition object collections into sets, typically in linear time.
- Using constraints to continually restrict sets until the desired solution sets are obtained.
- Performing top-down analysis to effectively direct the parsing process.

---

[3] One of the most important characteristic lengths in diagrams is the height (font size) of the smallest text. Anything of that size is considered small or short and anything that is many times that size is large or long. This follows naturally from the standard conventions that people use for constructing diagrams — text is typically made as small as possible subject to the constraint that it be clearly legible. The other characteristic length is the width of the page or column in which the graphic appears.

- Allowing objects to participate in more than one structure, e.g., a shared wall between two rooms. This is a natural consequence of the constraint approach.
- Building spatially associative indexes of all primitives and derived objects to aid searching and the computation of relations.

### Efficiency and Spatial Indexing

All the primitives in a diagram are initially entered into a spatial index. In the example below, the index allows us to find just those graphical items that touch a given line, and a set of horizontally aligned items, in a time independent of the total diagram size. Each cell in the index array corresponds to a square region in the Cartesian space of the diagram ($2^{13}$ x $2^{13}$). The finest resolution used in the index is typically a 64 x 64 array of cells. In addition, a pyramid containing all smaller arrays of size $2^n$ x $2^n$ (n<6) is built with each smaller array covering the same total area at a coarser resolution. This allows us to quickly discover objects that are more distant from one another or objects that satisfy some relation with a lower level of precision. For each array of X,Y cells, single X and Y projection arrays are created in which each cell contains the projection (union) of all objects installed in that portion of the one-dimensional space. The X and Y spatial indexes allow horizontally and vertically aligned sets of objects to be found rapidly. Inverse indexes for all objects are also pre-computed, mapping from objects to the set of cells containing them.

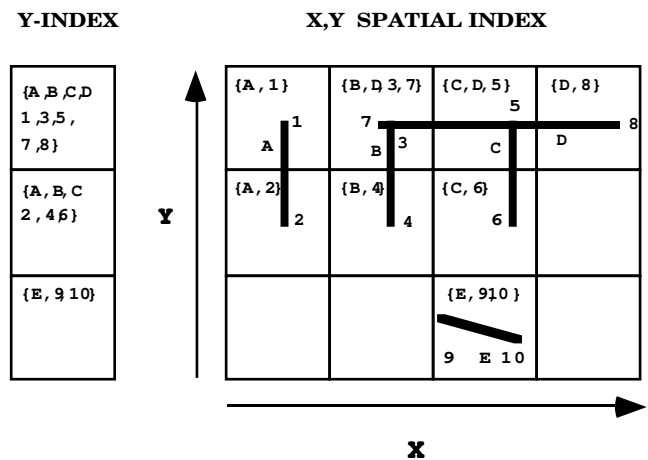

**Figure 3.** Spatial indexing of graphic primitives. A small portion of the much larger 64 x 64 array is shown. The vertical lines A, B and C, the horizontal line D and the diagonal line E are shown installed in the X,Y spatial index and the Y spatial index. (X index not shown). The line endpoints, 1-10 are also installed. The bracketed sets, {....}, are the items that are contained in or pass through a cell. The Y index set in a cell is the union of all cell contents to its right. This figure has N=5 for the five source lines, but an additional 10 endpoints are installed before parsing for use in alignment computations.

The example of Fig. 3 can be used to understand some of the details of parsing using grammar G1. The only X-Line in Fig. 3 found by Rule 2 is line D. "(touch X-Line ?)" generates the context value by using the inverse index from the line D to obtain the three cells it occupies. Then the set of all objects in those cells, the lines B, C and D and the endpoints 3, 5, 7 and 8 becomes the context value that is passed to the Ticks rule. They all "touch" the X-Line. The Ticks set rule first filters out all but Lines, the type specified for the set elements, leaving only B, C and D. It then filters out all but vertical Lines, leaving B and C and restricts to short Lines, still leaving B and C. It then checks to see if B and C are horizontally aligned, which they are, by seeing if their endpoints, e.g., their upper endpoints, 3 and 5, are contained in the same Y index cell, which they are. Then a Ticks set solution object containing B and C is returned to Rule 1 where, in this case, it is rejected because it only has two elements.

As parsing proceeds, higher-level objects are also installed in the spatial index. Thus, references to the Ticks object in Fig. 3, would be placed in the cells occupied by the lines B and C. This can be done more efficiently than the original installation, since the set of cells occupied by B and C are immediately available from the inverse index from objects to cells.

The spatial index can be used to generate or filter objects. All objects within some distance of an object O can be generated by looking in the cells occupied by O, at any chosen level of the spatial index pyramid. If a large context is passed to a rule, it can be filtered by generating a set obeying a constraint and intersecting it with the context. The spatial index can also be used to rapidly find all objects that are left, right, above or below a given object. For example, a search for data points in a graph can be done among objects that are above the x scale line and right of the y scale line. To find all objects to the right of some point P, the X index pyramid is searched from its root. The computation requires performing the union of the contents of at most n cells, where n is the depth of the pyramid, e.g., n=6. The union computation is linear in the total number of the (not necessarily distinct) objects in the n cells.

**A large grammar for data graphs**

The data graph of Fig. 1 was parsed using grammar G2, given in the Appendix. Much of grammar G2 is similar in form to G1, but it includes additional constructs.

Rule X-Line has the ":additional-slots" construct which specifies that an additional attribute be added to the LHS, in this case, "left-endpoint" which is bound to the left endpoint of the Line. Rule Y-Axis has the ":null" construct that allows the rule to be satisfied with null values for any or all of the constituents listed, if they cannot be found. Rule Y-Labels contains the ":largest" construct which forces this set rule to return only a single solution, the maximal set with the largest number of elements. The below constraint in the X-Axis rule contains the keyword ":strip", which causes it to ignore all objects that are left or right of the X-Axis-Line.

Basic functions on non-primitives are supported, e.g., *near*, *aligned*, *above*. But more specialized functions such as *a-length* for the high-level Data-Line object in Rule Data-Lines must be written with some knowledge of the structure of the object (e.g., a set of connected lines or curves).

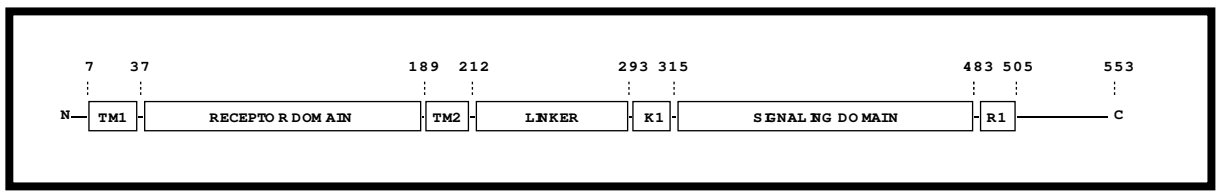

**Figure 4.** A gene diagram (N = 35) parsed with a grammar of eight rules. Parsing required 24 sec plus 14 sec for precomputing the spatial index.

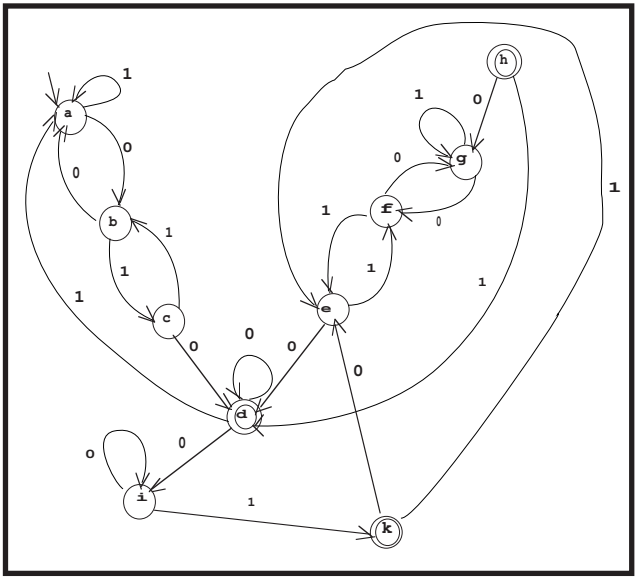

**Figure 5.** A finite-state automata sketch (FSA), N=124. Parsing required 65 sec plus 25 sec for precomputing the spatial index. The fact that the arc ends and arrowheads were drawn roughly and did not line up accurately posed no problems for the parser because the near constraint was used. Simple postprocessing of the parse gives the entire state-transition table so the FSA can be run. The arrows are recognized from their constituent lines, rather than assumed as primitives.

**Other approaches**

There are a number of systems that perform interpretation of engineering drawings, circuit diagrams and maps [6, 7].[4] Such systems usually start from a scanned image and try to create a high level description of the document. In general, those systems use complex domain-specific knowledge representations, making it difficult to apply the methods to different domains. In [8] a system for interpretation of large-scale hand-drawn logic circuit diagrams is presented. Symbols are recognized by a combination of feature extraction and pattern matching techniques, while decision trees and heuristics control the analysis process. The diagrams processed can be composed of up to 400 symbols and experimental results have shown that the recognition accuracy is about 95%. Other similar systems are the REDRAW [9] and GTX 5000 [10] that provide a more general framework for document analysis. The CIPLAN [11] is a system for interpretation maps that uses specialized domain knowledge at the pixel and higher levels of analysis. It implements a procedural network that associates entities with specific procedures for their identification. CIPLAN works in the domain of French city maps (plats). The application of CIPLAN to plats from other countries was not done because it required the integration of new specifications into the structure model. This is a typical situation in which efficiency is traded for adaptability. There is also substantial research in the low level aspects of document analysis [12, 13]. It mainly deals with segmentation, vectorization and feature extraction.

---

[4] While these have been popular application domains, the number of diagrams in the world's published technical literature far outstrips these, numbering over ten million a year. Our focus is on the biology literature, the largest single literature of the sciences, and one that contains a large proportion of diagrams.

Constraint-based grammars such as we use are recognized as useful for expressing relations among graphical objects in 2-dimensional space [3]. Others include Relation Grammars [14, 15], Graphical F-PATR Grammars [16], Picture Layout Grammars [17-19] and Constraint Set Grammars [20]. Despite the theoretical foundation that these approaches provide, it is not clear how to use them to achieve efficient parsing for particular application domains. Most are based on exhaustive bottom-up analysis and as a result, they are inefficient, especially in cases of many local matches that are not part of a complete solution. Wittenburg has developed a bottom-up tabular parsing algorithm that has successfully applied in the some interactive domains: flowchart and mathematical expressions interpretation [16], and document design [21]. In [19] an algorithm for spatial parsing is presented that is based on the CYK parsing algorithm. All of these approaches appear to have difficulty parsing diagrams of realistic size. RG/1 grammars were introduced [22] in order to make parsing tractable, but the grammars were then not expressive enough to deal with real diagrams. Wittenburg only gives very small examples in his various publications, so the question of parsing realistic diagrams remains open. Golin's thesis [19] gives some of the most detailed information on parsing using constraint grammars, but he concludes (Sec. 8.1.2) that the system is too slow to be of practical use. Helm gives examples from a number of domains, with N up to about 30, but no large diagrams are shown and no performance figures are given.

**Discussion**

The most important point to be made about the other work we have reviewed, is that it divides into two classes. In the first class are domain-specific systems that are capable of analyzing complex diagrams. But those systems typically have a lot of domain-specific code that can only be retargeted to another domain with great effort. In the second class are the more grammar-based approaches. These pay a lot of attention to proving formal properties of their grammars or fitting them into a rigorous language parsing framework. They are capable of describing a variety of domains. But the grammar-based systems do not appear to be efficient enough to parse diagrams of any really complexity, e.g., N=100 to 200 elements. Our system has analyzed over twenty diagrams from the published biology literature, average N=120.

The approach we have presented in this paper combines the flexibility of domain retargeting by writing alternate grammars, with the efficiency of the more domain-specific systems, something we believe has not been achieved before. Our system appears to succeed because of four unique features:

1. Spatial indexing
2. Sets as a grammar data type
3. Equivalence relations leading to maximal sets
4. Successive restriction of contexts through top-down analysis

Feature 1 has been described in detail. Features 2 and 3 work together. Feature 3 refers to the fact that relations such as near and aligned are approximate equivalence relations, which we call *Generalized Equivalence Relations* (GERs) [1], e.g., the relation *coincident* is a true equivalence relation, and *near* is a generalization of it. Diagrams are typically drawn with many items that are equivalent in some way, e.g., rectangles of the same size, arrowheads that are identical in size, data points that are the same shape and size, tick marks and their numerical labels that are aligned. This organization makes it simpler to draw and understand a diagram — it is tuned to the visual perceptual abilities that are innate in humans. Our system is designed to take advantage of the standard and natural paradigms used in diagram design.

In our top-down parsing strategy, the context is passed down as an inherited attribute and additional slots can be passed up as synthesized attributes [23]. Also, bounding boxes for complex objects are synthesized from constituents.

In conventional parsing, once a constituent is assigned a role, it is excluded from consideration for other roles — it cannot be "shared". But in graphics, sharing is common and our system handles sharing. For example, the analysis of Fig. 1 produces a solution with four parts in which each axis label, "Time After ..." and "Fraction ..." applies to all four graphs and the numeric tick labels apply to the two graphs above or to the right.

The system is written in Macintosh Common Lisp and uses the CLOS object system extensively. The grammar is preprocessed to discover all high-level objects and CLOS classes are created for them, including any additional slots specified in the rules. The Common Lisp macro facility makes it very easy to map declarative rules onto the needed constructors. A visual inspector, DUSI (Diagram Understanding System Inspector), has been implemented for development purposes that will highlight, in color, any CLOS graphic object in the display and conversely, locate the CLOS object corresponding to any displayed item.

# APPENDIX — Grammar G2 for data graphs

```
XY-Data-Graph -> Axis   X-Axis   Y-Axis   Data
        (Axis)
        (X-Axis   Axis)
        (Y-Axis   Axis)
        (Data     (contain Axis ?));

Axis -> X-Line   Y-Line
        (X-Line)
        (Y-Line  (touch  (left-endpoint X-Line) ?)
             :constraints
             (< (distance (left-endpoint X-Line) (bottom-endpoint Y-Line)) *tiny*));

X-Line -> Line
        (:additional-slots   (left-endpoint (left-endpoint (Line self))))
        (:constraints        (horizp Line)   (long Line));

Y-Line -> Line
        (:additional-slots   (bottom-endpoint  (bottom-endpoint (Line self))))
        (:constraints        (vertp Line)    (long Line));

        ***********    <X-AXIS > ****************
X-Axis -> X-Axis-Line   X-Ticks   X-Labels   X-Text
        (:null X-Text)
        (X-Axis-Line  (X-Line context))
        (X-Ticks      (touch  X-Axis-Line ?)
            :constraints (>= (size X-Ticks) 2)   (above X-Ticks X-Axis-Line))
        (X-Labels     (below ? X-Axis-Line :strip t))
        (X-Text       (below-nearest ? X-Labels));

X-Ticks -> Set ( Line )
        (:element-constraints  (vertp Line)   (short Line))
        (:constraint horiz-aligned));

X-Labels -> Set ( Text )
        (:element-constraints  (horizp Text)   (numeric-textp Text))
        (:constraint horiz-aligned)
        (:largest t);

X-Text -> Set ( Text )
        (:element-constraints (horizp Text))
        (:largest t);

        ***********    < Y-AXIS > ****************
Y-Axis -> Y-Axis-Line   Y-Ticks   Y-Labels   Y-Text
        (:null Y-Ticks Y-Labels Y-Text)
        (Y-Axis-Line       (Y-Line context))
        (Y-Ticks           (touch Y-Axis-Line ?)
            :constraints (right Y-Ticks Y-Axis-Line))
        (Y-Labels          (left ? Y-Axis-Line :strip t))
        (Y-text            (left-nearest ? (or Y-Labels  Y-Axis-Line)));

Y-Ticks -> Set ( Line )
        (:element-constraints  (horizp Line)   (short Line))
```

```
        (:constraint vert-aligned);

Y-Labels -> Set ( Text )
        (:element-constraints (horizp Text) (numeric-textp Text))
        (:constraint vert-aligned)
        (:largest t);

Y-Text -> Set ( Text )
        (:element-constraints (vertp Text))
        (:largest t);

      *********** < DATA >  ****************
Data -> Data-Lines  Data-points ;

Data-Lines  -> set ( Data-Line )
        (:element-constraints  (> (a-length Data-Line) *very-long*));

Data-Line  -> set ( Line )
        (:constraint connected);

Data-Line  -> set ( Curve )
         (:constraint connected);

Data-Points -> set ( Data-Cluster );

Data-Cluster -> set ( Data-Point )
        (:constraint same-type);

Data-Point -> Circle ;

Data-Point -> Polygon
        (:constraints  (rectanglep Polygon) (small Polygon));
```